\documentclass[sigconf, nonacm]{acmart}
\usepackage{textalpha}
\AtBeginDocument{%
  \providecommand\BibTeX{{%
    \normalfont B\kern-0.5em{\scshape i\kern-0.25em b}\kern-0.8em\TeX}}}

\usepackage{annotate-equations}
\usepackage{xcolor}
\usepackage{algorithm}
\usepackage{algpseudocode}

\definecolor{ForestGreen}{HTML}{009B55}
\definecolor{MidnightBlue}{HTML}{006795}

\usepackage{tabularx, environ}

\makeatletter

\newcolumntype{\expand}{}
\long\@namedef{NC@rewrite@\string\expand}{\expandafter\NC@find}

\NewEnviron{problem}[2][]{%
  \def\problem@arg{#1}%
  \def\problem@framed{framed}%
  \def\problem@lined{lined}%
  \def\problem@doublelined{doublelined}%
  \ifx\problem@arg\@empty%
    \def\problem@hline{}%
  \else%
    \ifx\problem@arg\problem@doublelined%
      \def\problem@hline{\hline\hline}%
    \else%
      \def\problem@hline{\hline}%
    \fi%
  \fi%
  \ifx\problem@arg\problem@framed%
    \def\problem@tablelayout{|>{\bfseries}lX|c}%
    \def\problem@title{\multicolumn{2}{|%
      >{\raisebox{-\fboxsep}}%
      p{\dimexpr\columnwidth-4\fboxsep-2\arrayrulewidth\relax\rule{0pt}{4ex}\hspace{-1em}}%
      |}{%
        \textsc{\Large #2}%
      }}%
  \else
    \def\problem@tablelayout{>{\bfseries}lXc}%
    \def\problem@title{\multicolumn{2}{>%
      {\raisebox{-\fboxsep}}%
      p{\dimexpr\columnwidth-4\fboxsep\relax\rule{0pt}{3ex}\hspace{-1em}}%
      }{%
        \textsc{\Large #2}%
      }}%
  \fi%
  \bigskip\par\noindent%
  \begin{tabularx}{\columnwidth}{\expand\problem@tablelayout}%
    \problem@hline%
    \problem@title\\[2\fboxsep]%
    \BODY\\\problem@hline%
  \end{tabularx}%
  \medskip\par%
}
\makeatother

\begin{document}

\title{Calculating lexicase selection probabilities is NP-Hard}


\author{Emily Dolson}
 \email{dolsonem@msu.edu}
 \orcid{0000-0001-8616-4898}
 \affiliation{%
   \institution{Michigan State University}
   \city{East Lansing}
   \state{Michigan}
   \country{USA}
 }

\renewcommand{\shortauthors}{Dolson}

\begin{abstract}
Calculating the probability of an individual solution being selected under lexicase selection is an important problem in attempts to develop a deeper theoretical understanding of lexicase selection, a state-of-the art parent selection algorithm in evolutionary computation. Discovering a fast solution to this problem would also have implications for efforts to develop practical improvements to lexicase selection. Here, I prove that this problem, which I name {\sc lex-prob}, is $NP$-Hard. I achieve this proof by reducing {\sc SAT}, a well-known $NP$-Complete problem, to {\sc lex-prob} in polynomial time. This reduction involves an intermediate step in which a popular variant of lexicase selection, $\epsilon$-lexicase selection, is reduced to standard lexicase selection.
This proof has important practical implications for anyone needing a fast way of calculating the probabilities of individual solutions being selected under lexicase selection. Doing so in polynomial time would be incredibly challenging, if not all-together impossible. Thus, finding approximation algorithms or practical optimizations for speeding up the brute-force solution is likely more worthwhile. 
This result also has deeper theoretical implications about the relationship between $\epsilon$-lexicase selection and lexicase selection and the relationship between {\sc lex-prob} and other $NP$-Hard problems.
%
%
%
\end{abstract}



\maketitle

\section{Introduction}



Lexicase selection is a state-of-the-art technique for selecting parents to reproduce in genetic programming \citep{spector_assessment_2012, helmuth_general_2015, helmuth_benchmarking_2020}. It is also increasingly being successfully applied in evolutionary computation domains outside of genetic programming \citep{metevier_lexicase_2019, lalejini_artificial_2022}. Consequently, it is worth developing a more rigorous theoretical understanding of lexicase selection, what makes it so effective, and what its limitations are. Most theoretical analyses of lexicase selection to date have involved calculating the probability of a given individual being selected on a given selection event \citep{la_cava_probabilistic_2018, dolson_ecological_2018}. This value, called $P_{lex}$ in \citep{la_cava_probabilistic_2018}, is critical to determining the long-term outcome of lexicase selection. 

This analysis is impeded by the fact that the naive approach to calculating $P_{lex}$ has exponential run time \citep{la_cava_probabilistic_2018}. Various optimizations can be applied to dramatically improve this run-time in practice, which makes it tempting to believe that calculating $P_{lex}$ in polynomial time might be possible. The creation of a polynomial-time algorithm for calculating $P_{lex}$ would facilitate more advanced theoretical analyses and potentially open up new avenues for refining lexicase selection. For example, \citep{la_cava_probabilistic_2018} points out that it would be tempting to plug $P_{lex}$ into a roulette selection algorithm as a potential simplification to the algorithm if calculating it were not so expensive in terms of run time.

So far, no one has succeeded in developing a polynomial-time algorithm for calculating $P_{lex}$. This fact raises the following question: is attempting to do so likely to pay off? Here, I provide evidence that it is not, by proving that the problem of calculating $P_{lex}$ is $NP$-Hard. 
These results extend to calculating $P_{lex}$ for other lexicase selection variants such as $\epsilon$-lexicase \citep{la_cava_epsilon-lexicase_2016}, downsampled lexicase \citep{hernandez_random_2019,ferguson_characterizing_2020}, and others \citep{spector_relaxations_2018}.

\section{Background}

\subsection{Lexicase Selection}

Lexicase selection is designed to operate in contexts in which candidate solutions in a population are evaluated on multiple fitness criteria. In genetic programming, for example, these fitness criteria are often test cases that a program may be evaluated on. It is common to use a large number of fitness criteria (\textit{e.g.} hundreds). Rather than combining a candidate solution's performance on each criterion into a single score, lexicase selection considers each one individually. 

To select a candidate solution to reproduce under lexicase selection, a random fitness criterion is chosen. All but the best-performing candidate solutions on that criterion are eliminated from consideration for selection. Then another random fitness criterion is chosen. Out of the set of solutions that were still in contention for selection, all but the best performers on this new criterion are then eliminated. This process continues until either only one candidate solution remains, at which point that candidate is chosen to reproduce. If all fitness criteria have been used and there are still multiple candidate solutions remaining, one of them is selected randomly. For a more in-depth explanation with examples, see \citep{spector_assessment_2012, la_cava_probabilistic_2018}.

\begin{algorithm}
\caption{The Lexicase selection algorithm}\label{alg:cap}
\begin{algorithmic}
\Require P - a vector of $N$ candidate solutions, each represented as a vector of scores on $M$ fitness criteria 
\Ensure A single candidate solution from P that should reproduce
\State $F \gets 1...M$ \Comment{Set of fitness criteria to consider}
\State $S \gets P$ \Comment{Set of candidate solutions to consider}
\While{$|P| > 1$ \& $|F| > 0$}
\State curr $\gets$ random criterion in $F$
\State best $\gets$ -1 \Comment{Assumes fitness is being maximized}
\For{solution : $S$}
\If{solution[curr] > best}
\State best $\gets$ solution[curr]
\EndIf
\EndFor
\State $S \gets$ all members of $S$ with score best on criterion curr
\EndWhile

\hspace{-2em}\Return{A random candidate solution in $S$}
\end{algorithmic}
\end{algorithm}

\subsection{\texorpdfstring{$\epsilon$-L} Lexicase Selection}

A number of variations on the core concept of lexicase selection have been developed. The most relevant to this paper is $\epsilon$-Lexicase Selection, which is commonly used when fitness criteria are floating point numbers \citep{la_cava_epsilon-lexicase_2016}. This variation adds a parameter, $\epsilon$, which provides a margin for error in the identification of the top performers on a given fitness criterion. In traditional lexicase selection, candidate solutions must be tied for best on the current fitness criterion in order to remain in contention for selection. In $\epsilon$-lexicase selection, however, candidate solutions need only be within $\epsilon$ of the best score on the current fitness criterion to remain in contention.


\section{Preliminaries}

\subsection{Problem Definitions}

There are five problems that it is important to formally define for the purposes of the following proofs. The first is the problem of calculating selection probabilities under lexicase selection ($P_{lex}$). I will call this problem the {\sc Lexicase Selection Probabilities Problem}, and will abbreviate it as {\sc lex-prob}. It is formally defined as follows:

\begin{problem}[ruled]{Lexicase Selection Probabilities Problem (lex-prob)}
  Input: & 1) a two-dimensional vector, $P$, containing $N$ vectors of length $M$. Each of the $N$ vectors represents a candidate solution in the population, and the values in the vector represent that solution's scores on each of the $M$ fitness criteria; 2) an integer $i$ indicating the index of a vector in $P$ \\
  Output: & The probability that vector $i$ would be selected from population $P$ by lexicase selection ($P_{lex_i}$)\\
\end{problem}

Note that this definition applies to calculating the selection probability for a single member of the population. In many cases, it is desirable to calculate selection probabilities for all members of a population. In practice, there are substantial efficiency gains from calculating probabilities for all members of the population at the same time. However, because the problem of calculating the probability of selection for a single member of the population is strictly easier, I will focus on it; proving that calculating $P_{lex}$ for even a single member of the population is $NP$-Hard will trivially show that calculating $P_{lex}$ for the entire population is also $NP$-Hard. Indeed, both versions of this problem can be reduced to each other in polynomial time\footnote{Calculating $P_{lex}$ for the entire population can be achieved by calculating $P_{lex}$ for each individual member of the population (i.e. $N$ times). Conversely, calculating $P_{lex}$ for an individual can be achieved by calculating it for the whole population and examining the value for that individual.} and thus they must be in the same computational complexity class. 

When proving that problems are $NP$-Hard, it is often easier to work with the \textit{decision} version of a problem, i.e. a version of the problem that returns a Boolean True or False as an answer. I define the decision version of {\sc lex-prob} as follows: 

\begin{problem}[ruled]{Decision Version of lex-prob}
  Input: & same as {\sc lex-prob}\\
  Output: & A Boolean value indicating whether there is a non-zero probability that vector $i$ would be selected from population $P$ by lexicase selection ($P_{lex_i}$)\\
\end{problem}

Note that, as is typically the case with decision versions of problems, the decision version of {\sc lex-prob} trivially reduces to the full version; output from the full version can trivially be converted to output for the decision version by comparing the returned $P_{lex}$ value to 0. Consequently, the full version can only be harder than the decision version, and so proving the decision version to be $NP$-Hard also proves the full version to be $NP$-Hard.

Note also that there is not an obvious way to reduce the full versions of this problem back to the decision version. This observation implies that it is possible that the decision versions is legitimately an easier problem than the full version. While this observation does not affect our proof, it is mildly interesting as it contrasts with many other problems (e.g. optimization problems) for which the reduction can be done in both directions.

We will also use the $\epsilon$-Lexicase Selection Probabilities Problem (abbreviated as {\sc $\epsilon$-lex-prob}) as a step in our proof. {\sc $\epsilon$-lex-prob} is the problem of calculating $P_{lex}$ under $\epsilon$ lexicase selection. It is formally defined as follows:

\begin{problem}[ruled]{$\epsilon$-Lexicase Selection Probabilities Problem ($\epsilon$-lex-prob)}
  Input: & All of the inputs to {\sc lex-prob}, plus a floating point value, $\epsilon$, indicating the allowed margin of error for selecting elite candidate solutions on each fitness criterion\\
  Output: & The probability that vector $i$ would be selected from population $P$ by $\epsilon$-lexicase selection ($P_{\epsilon-lex_i}$) using the specific $\epsilon$ value\\
\end{problem}

As with {\sc lex-prob}, I will primarily use the decision version of {\sc $\epsilon$-lex-prob} in our proof. Like {\sc lex-prob}, {\sc $\epsilon$-lex-prob} can trivially be reduced to its decision version (but not the other way around), meaning the full {\sc $\epsilon$-lex-prob} problem must be as hard as or harder than its decision version. I formally define the decision version of {\sc $\epsilon$-lex-prob} as follows:

\begin{problem}[ruled]{Decision version of $\epsilon$-lex-prob}
  Input: & same s {\sc $\epsilon$-lex-prob}\\
  Output: & A Boolean value indicating whether there is a non-zero probability that vector $i$ would be selected from population $P$ by $\epsilon$-lexicase selection ($P_{\epsilon-lex_i}$) using the specific $\epsilon$ value\\
\end{problem}

To prove that {\sc lex-prob} is $NP$-Hard, one must show that some problem that is already known to be $NP$-Hard can be reduced to {\sc lex-prob} in polynomial time. Thus, a known $NP$-Hard problem must be selected to use as a starting point. Here, I will use the classic $NP$-Hard problem, {\sc Boolean Satisifiability} (commonly abbreviated as {\sc SAT}) \citep{miller_reducibility_1972}, which is defined as follows:

\begin{problem}[ruled]{Boolean Satisfiability (SAT)}
  Input: & 1) a set, $V$, of $n$ variables, and 2) a set, $C$, of $m$ clauses containing variables in $V$ or their negations \\
  Output: & A Boolean value indicating whether it is possible to assign truth values (i.e. True or False) to each variable in $V$ such that each clause in $C$ contains at least one element that evaluates to True. \\
\end{problem}

\subsection{Brief review of computational complexity theory}

For readers unfamiliar with the $NP$-Hard class of problems and how to prove that problems are in this class, I offer a capsule summary. 

In order to talk about how hard problems are, we place them into computational complexity classes. Many problems that we regularly encounter are in the class $NP$ (non-deterministically polynomial), which is the set of problems for which we can verify in polynomial time that a given solution is correct \citep{miller_reducibility_1972}. Some problems in $NP$ are ``easy'' in the sense that polynomial-time solutions for them have been found. Others problems in $NP$ seem to be harder; no one has yet found a polynomial-time solution for them. The hardest of these problems are $NP$-Hard problems: the class of problems that have been proved to be at least as hard as the hardest problems in $NP$ \citep{cook_complexity_1971, miller_reducibility_1972}. While we do not know for sure whether it is possible to solve these problems in polynomial time, experts overwhelmingly agree that it is unlikely to be possible \citep{gasarch_guest_2019}.  If calculating $P_{lex}$ is $NP$-Hard, that is useful to know because it suggests that efforts to find a polynomial time algorithm are unlikely to pay off.

To prove that a problem is $NP$-Hard, we must take a problem that we already know is $NP$-Hard and show that it reduces (in polynomial time) to the problem we're interested in.  Here, I will reduce {\sc SAT} to {\sc $\epsilon$-lex-prob}, \textit{i.e.} I will convert instances of {\sc SAT} to instances of {\sc $\epsilon$-lex-prob} that, when solved, will also provide the solution to the underlying {\sc SAT} instance. By showing that any instance of {\sc SAT} can be converted to an equivalent instance of {\sc $\epsilon$-lex-prob} in polynomial time, I will prove that {\sc $\epsilon$-lex-prob} is at least as hard as {\sc SAT}. If someone discovered a polynomial-time algorithm that solved {\sc $\epsilon$-lex-prob}, that would mean that we also have a polynomial-time algorithm for {\sc SAT} (and, indeed, all of the $NP$-Complete problems). I will subsequently show that {\sc $\epsilon$-lex-prob} reduces in polynomial time to {\sc lex-prob}, and thus {\sc lex-prob} is, by the same logic, also $NP$-Hard.

Some readers may be more familiar with the class $NP$-Complete. $NP$-Complete problems are problems that are $NP$-Hard and are also in the set $NP$. Typically, only the decision versions of problems are in $NP$ (and therefore capable of being $NP$-complete), as verifying that the precise values returned by non-decision problems is seldom easier than calculating them in the first place. While I will show that the decision versions of {\sc $\epsilon$-lex-prob} and {\sc lex-prob} are $NP$-Complete, the emphasis of this paper is on the full versions of the problems. As these problems are not in $NP$, I have chosen to emphasize that the full versions are $NP$-Hard, rather than that the decision versions are $NP$-Complete. Both of those statements are true, however.

\section{{\sc lex-prob} and {\sc $\epsilon$-lex-prob} are $NP$-Hard}

We will prove that {\sc lex-prob} and {\sc $\epsilon$-lex-prob} are $NP$-Hard via the following sequence of reductions:

\[
\text{{\sc SAT}} \leq_{P} \text{{\sc $\epsilon$-lex-prob} (decision)} \leq_{P} \text{{\sc lex-prob} (decision)}
\]

In other words, I will show that {\sc SAT}, a known $NP$-Hard problem, can be reduced in polynomial time to the {\sc $\epsilon$-lex-prob} problem. Subsequently, I will show that the {\sc $\epsilon$-lex-prob} problem can be reduced in polynomial time to the {\sc lex-prob} problem. For convenience, I will use the decision version of {\sc $\epsilon$-lex-prob} and {\sc lex-prob}. As the decision versions of each trivially reduce in polynomial time to the full versions, this series of reductions will prove that both versions of these problems are $NP$-Hard. Formally:

\[
\text{{\sc SAT}} \leq_{P} \text{{\sc $\epsilon$-lex-prob} (decision)} \leq_{P} \text{{\sc $\epsilon$-lex-prob}}
\]
\[
\text{and}
\]
\[
\text{{\sc $\epsilon$-lex-prob} (decision)} \leq_{P} \text{{\sc lex-prob} (decision)} \leq_{P} \text{{\sc lex-prob}}
\]

We will also prove that the decision versions of {\sc $\epsilon$-lex-prob} and {\sc lex-prob} are $NP$-Complete by showing that, in addition to being $NP$-Hard, these problems are in $NP$.

\subsection{The decision versions of {\sc lex-prob} and {\sc $\epsilon$-lex-prob} are in $NP$}

In order to show that the decision version of {\sc lex-prob} and {\sc $\epsilon$-lex-prob} are $NP$-Complete, it must first be shown that they are in $NP$.

\begin{theorem}
\label{np_theorem}
{\sc lex-prob} and {\sc $\epsilon$-lex-prob} are in the set $NP$.
\end{theorem}

In the formal definition of a decision problem, its only output is True or False. However, for the purposes of demonstrating membership in the class $NP$, it is also allowed to return additional information (a ``certificate'') for use in verifying that the True/False output is correct. 

\begin{proof}
    Let the certificate returned by {\sc lex-prob} or {\sc $\epsilon$-lex-prob} be a list, $L$, denoting an order in which fitness criteria could be selected that would cause candidate solution $i$ to be selected by lexicase selection or $\epsilon$-lexicase selection.

    The size of $L$ will be $\mathcal{O}(M)$, where $M$ is the number of fitness criteria.

    To verify that the returned answer is correct, we simply need to step through the fitness criteria in the order specified by $L$, removing individuals from the original population, $P$, according to the rules of lexicase selection. This process can be carried out in $\mathcal{O}(MN)$ time, where $N$ is the number of candidate solutions in the population.

    As $MN$ is a polynomial, this verification can be carried out in polynomial time and {\sc lex-prob} and {\sc $\epsilon$-lex-prob} are both in $NP$.
\end{proof}


\subsection{{\sc $\epsilon$-lex-prob} is $NP$-Hard}

\begin{figure*}
    \centering
    \includegraphics[width=\linewidth]{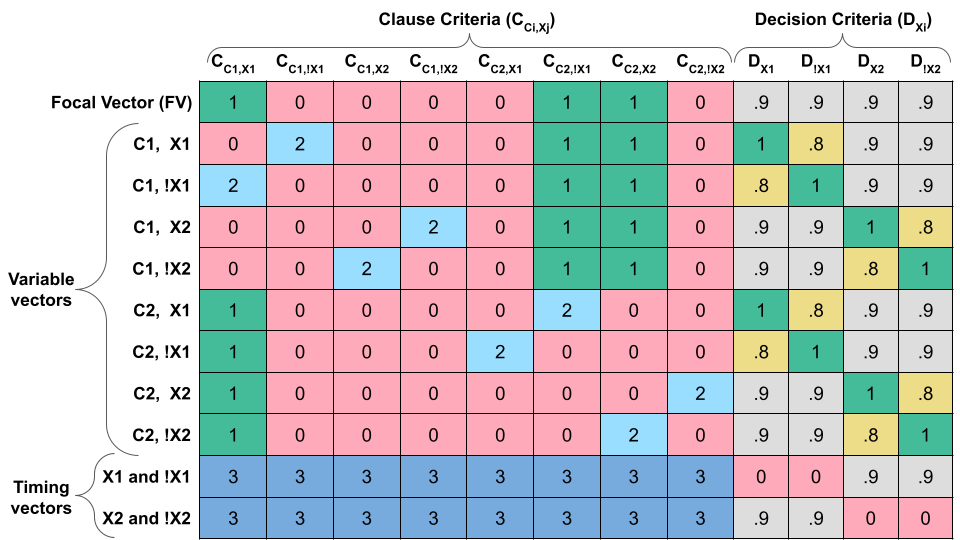}
    \caption{Worked example of reducing {\sc SAT} to {\sc $\epsilon$-lex-prob}. This is the {\sc $\epsilon$-lex-prob} instance corresponding to the {\sc SAT} following instance: $V = \{X1, X2\}$, $C = \{X1\}, \{!X1, X2\}$. Note that the focal vector has a non-zero probability of selection; it will win when, for example, fitness criteria are chosen in the following order: $D_{X1}$, $D_{X2}$, $C_{C1,X1}$, $C_{C2,X2}$ (at this point only the focal vector will remain in contention for selection and so we can skip the remaining eight fitness criteria). This sequence of fitness criteria corresponds to assigning $X1=true, X2=true$, which is indeed a set of assignments that satisfies the original instance of {\sc SAT}. }
    \label{fig:sat_to_eps}
\end{figure*}

We propose the following theorem:

\begin{theorem}
\label{elexicasetheorem}
{\sc $\epsilon$-lex-prob} is $NP$-Hard.
\end{theorem}

We will prove this theorem by reducing {\sc SAT}, a well-known $NP$-Complete problem \citep{miller_reducibility_1972}, to the decision version of {\sc $\epsilon$-lex-prob} in polynomial time. For a worked example of this reduction, see Figure \ref{fig:sat_to_eps}.

\begin{proof}
Given an instance of {\sc SAT} with $c$ clauses and $v$ variables, one can create an instance of {\sc $\epsilon$-lex-prob} with a population, $P$, containing $N = 1 + 2vc + v$ candidate solutions with scores on $M = 2vc + 2v$ fitness criteria. $P$ can be visualized as a matrix where row vectors are candidate solutions and column vectors are fitness criteria. $\epsilon$ will be set to .1. 

The first $2vc$ fitness criteria in each candidate solution correspond to the clauses in the original {\sc SAT} instance. The representation of each clause is made up of $2v$ fitness criteria, one for each variable and the negation of each variable. I will refer to these fitness criteria as $C_{C_{i},X_{j}}$, where $i$ is the index of the clause the fitness criterion is part of and $j$ is the id of the variable the fitness criterion refers to. For example, $C_{C2X3}$ would be the criterion corresponding to the value of the variable $X3$ in clause 2.

The next $2v$ fitness criteria are the ones that will be used to explore different assignments of truth values to variables in the {\sc SAT} instance. The order in which lexicase selection picks these fitness criteria will determine which variable assignments are made. I will refer to the position corresponding to choosing to set $Xi$ to true as $D_{Xi}$ (whereas the position that corresponds to setting it to false would be $D_{!Xi}$).

The first candidate solution in $P$, which I term the \textit{focal vector}, directly represents the {\sc SAT} instance. Its scores on the $C_{C_{i},X_{j}}$ fitness criteria will be set to indicate which variables are in each clause. If clause $i$ contains variable $j$, then $C_{C_{i},X_{j}}$ in the focal vector is set to 1. Otherwise, it is set to 0. All $D_{Xi}$ values are set to .9. The rules for creating the focal vector are stated formally in equation \ref{eq:focal_vector}.
\vspace{1.5em}

\begin{equation}
\label{eq:focal_vector}
\eqnmarkbox[violet]{fvi}{FV(x)} = \begin{cases} 
1 & \eqnmarkbox[teal]{ccixj}{x < 2cv} \text{ } \& \eqnmarkbox[blue]{var_xj}{X_{x\%(2v)}} \in \eqnmarkbox[magenta]{clause_ci}{C_{\lfloor i/(2v)\rfloor}} \\
0 & \eqnmarkbox[teal]{ccixj2}{i < 2cv} \text{ } \& \eqnmarkbox[blue]{var_xj2}{X_{x\%(2v)}} \notin \eqnmarkbox[magenta]{clause_ci2}{C_{\lfloor x/(2v)\rfloor}} \\
.9 & \eqnmarkbox[red]{dxi}{\text{otherwise}}
\end{cases}
\end{equation}

\annotate[yshift=1em]{left}{ccixj}{$x$ is a $C_{C_{i},X_{j}}$ criterion}
\annotate[yshift=1em]{right}{var_xj}{this criterion's variable ($X_j$) }
\annotate[yshift=-.2em]{below, right}{clause_ci2}{this criterion's \\ \sffamily\footnotesize clause ($C_i$)}
\annotate[yshift=-2.8em]{below,right}{fvi}{$x$th value of focal vector}
\annotate{below,right}{dxi}{$D_{Xi}$ objectives}
\vspace{1.5em}

The next $2vc$ candidate solutions in the population correspond to possible variable assignments for each clause. I will call these the \textit{variable vectors}. Let variable vector $VV_{ab}$ be the variable vector corresponding to clause $C_a$ and variable $X_b$. The $C_{C_{i},X_{j}}$ criteria values for $VV_{ab}$ are identical to the focal vector's, except for when $i=a$ (i.e. in the clause corresponding to $VV_{ab}$). $VV_{ab}$'s scores for all such fitness criteria ($C_{C_{a},X_{j}}$) will be 0, except when $j=inv(b)$. $VV_{ab}$ score for this fitness criterion, $C_{C_{a},!X_{b}}$, which corresponds to the negation of variable $X_b$ in clause $a$, will be set to 2. 

$D_{X_{i}}$ values for the variable vectors are set to .9, except in the position corresponding to that variable and its negation. For variable vectors corresponding to $X_{i}$, $D_{!X_{i}}$ will be set to .8, and $D_{X_{i}}$ will be set to 1. For variable vectors corresponding to $!X_{i}$, $D_{!X_{i}}$ will be set to 1, and $D_{X_{i}}$ will be set to .8. The rules for creating the focal vector are stated formally in equation \ref{eq:variable_vector}.

\vspace{3em}

\begin{equation}
\label{eq:variable_vector}
\eqnmarkbox[violet]{vvi}{VV_{xy}(z)} = \begin{cases} 
2 & \eqnmarkbox[teal]{ccixj2}{z < 2cv} \text{ } \& \eqnmarkbox[blue]{clausei}{x = {\lfloor z/(2v)\rfloor}}  \text{ } \& \eqnmarkbox[magenta]{inv_var}{\text{ inv}(y) = z\%(2v)}\\
0 & \eqnmarkbox[teal]{ccixj}{z < 2cv} \text{ } \& \eqnmarkbox[blue]{clausei2}{x = {\lfloor z/(2v)\rfloor}}  \text{ } \& \text{ inv}(y) \neq z\%(2v)\\
FV_z & \eqnmarkbox[teal]{ccixj}{z < 2cv} \text{ } \& x \neq {\lfloor z/(2v)\rfloor} \\
1 & \eqnmarkbox[orange]{dxi1}{z \geq 2cv \text{ } \& \text{ } z - 2cv = y}  \\
.8 & \eqnmarkbox[gray]{dxi3}{z \geq 2cv \text{ } \& \text{ } z - 2cv = inv(y)}  \\
.9 & \eqnmarkbox[red]{dxi2}{\text{otherwise}}
\end{cases}
\end{equation}

\annotate[yshift=1em]{left}{ccixj2}{$k$ is a $C_{C_{i},X_{j}}$ criterion}
\annotate[yshift=.5em]{right}{clausei}{$z$th criterion's \\ \sffamily\footnotesize clause is $C_x$}
\annotate[yshift=3em]{left}{inv_var}{$z$th criterion's variable is $!X_y$ ($z$th criterion is $C_{Cx!Xy}$)}
\annotate[yshift=-5em]{below,right}{vvi}{$z$th value of variable vector for variable $X_y$ and clause $C_x$}
\annotate[xshift=8em,yshift=.15em]{right}{dxi1}{$D_{Xy}$ criterion}
\annotate[]{below,right}{dxi3}{$D_{!Xy}$ criterion}
\annotate[]{below,right}{dxi2}{other $D_{Xi}$ criteria}
\vspace{1.5em}

\vspace{1.5em}

Lastly, there are $v$ vectors that I will refer to as ``timing'' vectors. These vectors have all $C_{C_{i},X_{j}}$ positions set to 3. Each timing vector corresponds to a single variable and its negation. $D_{X_{i}}$ and $D_{!X_{i}}$ are set to 0 when $i$ is that vector's corresponding variable. Otherwise, they are set to .9.

\begin{equation}
    \label{eq:timing_vector}
    \eqnmarkbox[violet]{tvxy}{TV_{x}(y)} = \begin{cases}
        3 & \eqnmarkbox[teal]{ccixj}{y < 2cv} \\
        0 & \eqnmarkbox[orange]{dxi1}{y \geq 2cv \text{ } \& \text{ } (y - 2cv = x \text{ } | \text{ } y - 2cv = \text{inv}(x))}  \\
        .9 & \eqnmarkbox[red]{dxi2}{otherwise}
    \end{cases}
\end{equation}

\annotate[yshift=-3em]{below, right}{tvxy}{$y$th value of timing vector for variable $X_x$}
\annotate[yshift=1em]{above, left}{ccixj}{$y$ is a $C_{C_{i},X_{j}}$ criterion}
\annotate[]{below, right}{dxi2}{other $D_{Xi}$ criteria}
\annotate[yshift=1em]{above, right}{dxi1}{$D_{Xx}$ or $D_{!Xx}$}
\vspace{1.5em}

\vspace{1.5em}

The timing vectors insure that, before any of the $C_{C_{i},X_{j}}$ criteria are selected, values have to be assigned to all variables by selecting either $D_{X_{i}}$ or $D_{!X_{i}}$ for all $i$. Any ordering of fitness criteria in which a $C_{C_{i},X_{j}}$ position is selected prematurely will result in one of the timing vectors winning.

When $D_{X_{i}}$ is selected, it eliminates all variable vectors corresponding to $D_{!X_{i}}$, and visa versa, because .8 is less than 1 by more than $\epsilon$. The other candidate solutions, however, are not eliminated, as .9 is within $\epsilon$ of 1. Selecting $D_{X_{i}}$ or $D_{!X_{i}}$ also eliminates the timing vector corresponding to variable $i$, which has 0s in both positions. Selecting a variable after selecting its negation or the other way around will have no effect, as all remaining values for that position will be .8 or .9 and so will be within $\epsilon$ of each other and tie.

Once variable assignments have been made, $C_{C_{i},X_{j}}$s criteria can now safely be selected. The focal vector will only have a chance of winning if, for each clause, there is at least one position where the focal vector has a 1 and no variable vectors with 2s remain in contention.

The different objective orderings in lexicase selection can be thought of as a decision tree with $N!$ leaf nodes, each representing a different ordering of objectives. Each level of the tree represents choosing the next objective, and so each node at level $i$ has $N-i$ child nodes (since i objectives have already been selected, there are $N-i$ options remaining).

For the purposes of this reduction, I only care about the portions of this tree in which one of the $D_{X_{i}}$ positions for each variable is ordered before any of the $C_{C_{i},X_{j}}$ positions. These orderings represent trying all possible combinations of truth assignments to the variables. The focal vector will have a non-zero chance of selection if and only if one of those combinations of truth assignments corresponds to a truth assignment that satisfies the original {\sc SAT} instance. Therefore, this transformation is a valid reduction from {\sc SAT} to {\sc lex-prob}.

This reduction takes $\mathcal{O}((1 + 2vc + v)(2vc + 2v))$ time, because each candidate solution in $P$ can be constructed in linear time with respect to the number of fitness criteria that are being created. This time complexity simplifies to $\mathcal{O}(v^2c^2)$. Therefore, {\sc SAT} reduces in polynomial time to {\sc $\epsilon$-lex-prob}, meaning that {\sc $\epsilon$-lex-prob} must be at least as hard as {\sc SAT}. Since {\sc SAT} is $NP$-Complete, {\sc $\epsilon$-lex-prob} must be $NP$-Complete as well.

\end{proof}

\subsection{{\sc lex-prob} is NP-Hard}

\begin{figure}
    \centering
    \includegraphics[width=\linewidth]{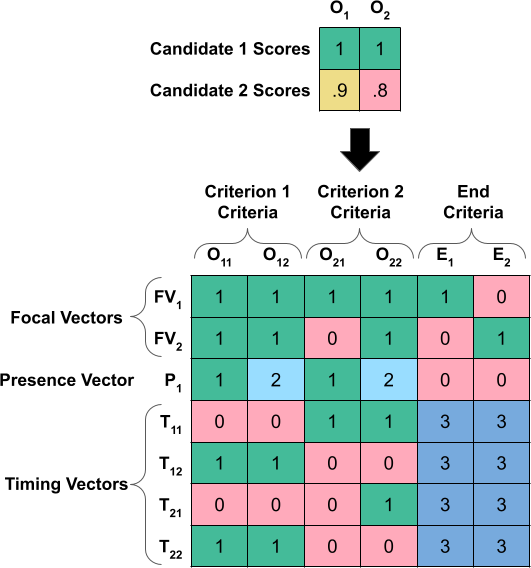}
    \caption{Worked example of reducing {\sc $\epsilon$-lex-prob} to {\sc lex-prob}. The top matrix shows the input set of population scores, $P$, for {\sc $\epsilon$-lex-prob}. The bottom matrix shows the corresponding set of population scores, $P'$ for {\sc lex-prob}. Note that focal vector 1 ($FV_1$) has a non-zero probability of selection; it will win, for example, when fitness criteria are selected in the following order: $O_{21}, O_{11}, E_{1}$ (at this point only $FV_1$ remains in contention for selection so we can stop choosing criteria). $FV_2$, on the other hand, will not be selected under any circumstances.}
    \label{fig:eps_to_lex}
\end{figure}

We have shown that {\sc $\epsilon$-lex-prob} is $NP$-Hard. However, that does not necessarily mean that {\sc lex-prob} is $NP$-Hard. Perhaps the inclusion of the $\epsilon$ parameter makes {\sc $\epsilon$-lex-prob} substantially harder. Here, I prove that {\sc lex-prob} is indeed $NP$-Hard as well. For a worked example of this reduction, see Figure \ref{fig:eps_to_lex}. 

\begin{theorem}
\label{elexicasetheorem}
{\sc lex-prob} is $NP$-Hard.
\end{theorem}

To prove that {\sc lex-prob} is also $NP$-Hard, I will reduce {\sc $\epsilon$-lex-prob}, which we now know to be $NP$-Hard, to {\sc lex-prob} in polynomial time.

\begin{proof}

Given an instance of {\sc $\epsilon$-lex-prob} with a population $P$ of $N$ candidate solutions, $M$ fitness criteria, some value $\epsilon$, and a candidate solution $i$ in $P$ to calculate $P_{lex}$ for, one can create an instance of {\sc lex-prob} by constructing a new population, $P'$, and returning $P_{lex}$ of its $i$th member. $P'$ will contain at most $2N + NM$ candidate solutions and $NM + N$ fitness criteria. 

The first $N$ candidate solutions will be referred to as focal vectors $i$ through $N$ and will correspond directly to the candidate solutions in $P$ ($P_1$ through $P_N$). $P'$ will be constructed such that $P_{lex}$ for a focal vector in $P'$ ($P'_i$) under lexicase selection is greater than 0 if and only if the corresponding candidate solution in $P$ ($P_i$) under epsilon lexicase selection also had $P_{lex}$ greater than 0. 

The key property of $\epsilon$-lexicase that makes it meaningfully different from standard lexicase selection is that in $\epsilon$-lexicase a variety of sets of ties are possible within the same fitness criterion, depending on which candidate solutions have yet to be eliminated. For example, consider three candidate solutions, A, B, and C, which respectively have scores 1, .9, and .8 on some fitness criterion. If $\epsilon=.1$ and A has not yet been eliminated when this criterion is chosen, A and B will tie, remaining in contention while C is eliminated. However, if A has been eliminated when this criterion is chosen, B and C will tie and both remain in contention for selection. 

To reproduce this capability, for each fitness criterion in $P$, there are $N$ fitness criteria in $P'$ (for a total of $NM$ fitness criteria). I will refer to these fitness criteria as $O_{ij}$, where $i$ is the fitness criterion it corresponds to in $P$, and $j$ refers to the id of this objective among the other objectives that also correspond to $i$. The $O_{ij}$ criteria will be used to reflect different sets of candidate solutions that could have tied on criterion $O_i$ under various circumstances. Specifically, candidate solutions will have $O_{ij}$ scores of 1 if their score on $O_i$ in $P$ is greater than or equal to the $j$th highest score that any candidate solution had on criterion $O_i$ in $P$ minus $\epsilon$.

However, having multiple new fitness criteria associated with each original fitness criterion causes problems if all of the new fitness criteria eventually need to be selected. For this reason, there are $N$ additional fitness criteria that effectively short circuit the process of selecting fitness criteria and ``end'' selection early. I will call these the $E_i$ criteria. Each focal vector will have a score of 1 on one $E_i$ criterion and scores of 0 on all others. Thus, selecting any of the $E_i$ criteria will immediately cause candidate solution $i$ to win.

Formally, the scores for each focal vector can be described by the following equation:

Let $S_i$ be a reverse-sorted vector of all scores in $P$ on fitness criterion $i$.

\vspace{3em}

\begin{equation}
\eqnmarkbox[violet]{fvxy}{FV_{x}(y)} = \begin{cases} 
                1 & \eqnmarkbox[red]{oij}{y < NM} \text{ }  \& \text{ }  \eqnmarkbox[orange]{px}{P_{x,\lfloor y/N \rfloor}} \geq \eqnmarkbox[magenta]{s}{S_{\lfloor y/N \rfloor, y\%N}} - \eqnmarkbox[gray]{eps}{\epsilon} \\
                0 & \eqnmarkbox[red]{oij2}{y < NM} \text{ }  \& \text{ }  \eqnmarkbox[orange]{px1}{P_{x,\lfloor y/N \rfloor}} < \eqnmarkbox[magenta]{s2}{S_{\lfloor y/N \rfloor, y\%N}} - \eqnmarkbox[gray]{eps2}{\epsilon} \\
                1 & \eqnmarkbox[blue]{e}{y \geq NM \text{ }\text{ }  \& \text{ }  y - NM = x} \\
                0 & \eqnmarkbox[teal]{e2}{\text{otherwise}}
            \end{cases}
\end{equation}
\annotate[yshift=1em]{above,left}{oij}{criterion $y$ is an $O_{ij}$ criterion}
\annotate[]{below,right}{e}{criterion $y$ is $E_{x}$}
\annotate[]{below,right}{e2}{criterion $y$ is a \\ \sffamily\footnotesize different $E_{i}$ criterion}
\annotate[yshift=2.5em]{above,left}{px}{$x$th member of $P$'s score \\ \sffamily\footnotesize on corresponding criterion ($O_{\lfloor y/N \rfloor}$)}
\annotate[yshift=1em]{above,right}{s}{reverse sorted \\ \sffamily\footnotesize vector of \\ \sffamily\footnotesize objective $i$ scores, \\ \sffamily\footnotesize $k$'th index}
\annotate[yshift=-1em]{below,left}{eps2}{from {\sc $\epsilon$-lex-prob} \\ \sffamily\footnotesize instance}
\annotate[yshift=-5em]{below, right}{fvxy}{$j$th value of $i$th focal vector}

\vspace{2em}

Ensure that the focal vectors can only be selected when appropriate requires adding two more sets of candidate solutions. The first of these are the \textit{presence vectors}. While $\epsilon$-lexicase selection allows multiple sets of ties to occur, the sets of ties that should be allowed to occur depend on the solutions still in contention for selection. Specifically, ties on lower scores on a criterion can only occur when candidate solutions with higher scores on that criterion are eliminated before that criterion is chosen. The presence vectors keep track of which solutions are still in contention and prevent focal vectors from being selected on the basis of $O_{ij}$ criteria representing ties that aren't available yet. Presence vectors each correspond to a focal vector that they are recording the presence of. They are identical to their corresponding focal vector, except that they have scores of 2 for any $O_{ij}$ criteria representing a tie that is only possible when that focal vector has been eliminated. Consequently, presence vectors will be eliminated from contention for selection under the same circumstances as their corresponding focal vector. However, they will also prevent any focal vectors from being selected using an ordering of fitness criteria that uses an $O_{ij}$ criterion before it is allowed. $P'$ will have at most $N$ presence vectors - one for every focal vector whose presence invalidates an $O_{ij}$ criterion.

\vspace{2em}

\begin{equation}
\eqnmarkbox[violet]{pvxy}{PV_{x}(y)} = \begin{cases} 
                2 & \eqnmarkbox[red]{oij}{y < NM} \text{ }  \& \text{ }  \eqnmarkbox[orange]{px}{P_{x,\lfloor y/N \rfloor}} > \eqnmarkbox[magenta]{s}{S_{\lfloor y/N \rfloor, y\%N}}  \\
                FV_x(y) & \eqnmarkbox[red]{oij1}{y < NM} \text{ }  \& \text{ }  \eqnmarkbox[orange]{px2}{P_{x,\lfloor y/N \rfloor}} \leq \eqnmarkbox[magenta]{s}{S_{\lfloor y/N \rfloor, y\%N}} \\
                0 & \eqnmarkbox[teal]{e}{\text{otherwise}}
            \end{cases}
\end{equation}
\annotate[yshift=1em]{above,right}{oij}{criterion $y$ is \\ \sffamily\footnotesize $O_{ij}$ criterion}
\annotate[]{below,right}{e}{all $E_{i}$ criteria are 0}
\annotate[yshift=1em]{above,right}{px}{score of $x$th member of $P$ \\ \sffamily\footnotesize on corresponding criterion \\ \sffamily\footnotesize ($O_{\lfloor y/N \rfloor}$)}
\annotate[yshift=-3em]{below,left}{s}{$(y\%N)$th value of a reverse-sorted \\ \sffamily\footnotesize vector of scores on ($O_{\lfloor y/N \rfloor}$) from $P$}
\annotate[yshift=3em]{above, right}{pvxy}{$y$th value of presence \\ \sffamily\footnotesize vector for $x$th focal vector}

\vspace{2em}

Finally, there are the \textit{timing vectors}. These candidate solutions ensure that at least one $O_{ij}$ criterion representing each of the fitness criteria in $P$ is selected before any of the $E_i$ fitness criteria are selected. There are $NM$ timing vectors, and each is associated with both a fitness criterion in $P$ and a focal vector in $P'$. Timing vectors are designed to be eliminated whenever 1) a corresponding focal vector is eliminated, or 2) a corresponding fitness criterion is chosen.

The timing vectors have scores of 3 for all $E_i$ fitness criteria, insuring that none of the focal vectors can win when an $E_i$ criterion is chosen too early. Each timing vector corresponds to a specific criterion, $O_i$ in $P$, and has a score of 0 for all $O_{ij}$ criteria corresponding to that criterion. All other scores in a timing vector are identical to the scores for its corresponding focal vector. Consequently, the only way to eliminate all timing vectors without also eliminating all focal vectors is to choose an $O_{ij}$ criterion for each value of $i$ (\textit{i.e.} an $O_{ij}$ criterion corresponding to each criterion in $P$).

Values for timing vectors are defined formally as follows:

\vspace{2em}

\begin{equation}
\eqnmarkbox[violet]{fvxy}{TV_{xy}(z)} = \begin{cases} 
                0 & \eqnmarkbox[red]{oij}{z < NM} \text{ }  \& \text{ }  \eqnmarkbox[orange]{px}{\lfloor z/N \rfloor} = x \\
                FV_y(z) & \eqnmarkbox[red]{oij2}{z < NM} \text{ }  \& \text{ }  \eqnmarkbox[orange]{px1}{\lfloor z/N \rfloor} \neq x \\
                3 & \eqnmarkbox[teal]{e2}{\text{otherwise}}
            \end{cases}
\end{equation}
\annotate[yshift=2em]{above,left}{oij}{criterion $z$ is an $O_{ij}$ criterion}
\annotate[]{below,right}{e2}{criterion $z$ is an $E_{i}$ criterion}
\annotate[yshift=1em]{above,right}{px}{id of criterion in $P$ that \\ \sffamily\footnotesize criterion $z$ corresponds to)}
\annotate[yshift=-3em]{below, right}{fvxy}{$z$th value of timing vector corresponding to the $x$th \\ \sffamily\footnotesize criterion in $P$ and the $y$th focal vector}

\vspace{4em}

This configuration ensures that, for every criterion $O_i$ in $P$, a corresponding objective $O_{ij}$ in $P'$ must be chosen before a focal vector can be selected. Further, it ensures that all ties that were possible in the {\sc $\epsilon$-lex-prob} instance are possible in our new {\sc lex-prob} instance, but only after the appropriate focal vectors have been eliminated. But what happens if a focal vector is eliminated and then we choose an $O_{ij}$ criterion representing a scenario where it had the highest fitness (\textit{i.e.} we choose $O_{ij}$ after $S_{ij}$ has been eliminated)? At this point, there are two possible scenarios: 1) all of the remaining scores on this $O_{ij}$ criterion are 0s (\textit{i.e.} all of the candidate solutions that were tied for best have been eliminated), or 2) there are some candidates still in contention with scores of 1 on this criterion. 

In case 1, the timing vector for this criterion ($O_i$) will not be eliminated as the focal vectors' scores of 0 will tie with timing vectors' scores of 0. Thus, another $O_{ij}$ criterion for this $i$ will need to be chosen before a focal vector can be selected. 

In case 2, the candidate solutions with non-zero scores would have remained in contention had this $O_{ij}$ criterion been chosen before whichever criterion eliminated $S_{ij}$. Thus, choosing that criterion and then $O_{ij}$ is functionally equivalent to having chosen these criteria in the opposite order; in either case, we get the set of candidate solutions that are elite on both criteria.

Therefore, this scenario does not actually pose a problem for this reduction.

This reduction ensures that the probability of selecting focal vector $i$ from $P'$ under lexicase selection will be non-zero if and only if the probability of selecting candidate solution $i$ from $P$ under $\epsilon$ lexicase selection was non-zero. This reduction takes $\mathcal{O}((2N + NM)(NM + N) + MNlog(N))$ time. Each of the $\mathcal{O}(2N + NM)$ new candidate solutions can be constructed in linear time with respect to the number of new fitness criteria ($NM + N$). Sorting the scores on each fitness criterion in $P$ will take an additional $\mathcal{O}(MNlog(N))$ time. This time complexity can be simplified to $\mathcal{O}(N^2M^2)$.

Consequently, any instance of {\sc $\epsilon$-lex-prob} can be solved using a solution to {\sc lex-prob} plus a polynomial time conversion. Thus, {\sc lex-prob} cannot be any easier than {\sc $\epsilon$-lex-prob}. As {\sc $\epsilon$-lex-prob} is $NP$-Hard, {\sc lex-prob} is also $NP$-Hard.

\end{proof}

\subsection{Extension to other lexicase selection variants} 

Given that {\sc lex-prob} is $NP$-Hard, calculating $P_{lex}$ under most other lexicase selection variants can trivially be proven to be $NP$-Hard as well. Often, lexicase selection variants add an additional parameter (or parameters) for which there exists some value such that the variant is equivalent to standard lexicase selection. One can trivially reduce lexicase selection to these variants by setting the parameters to those values. Consequently, their equivalents to {\sc lex-prob} are also $NP$-Hard, because any solution to them would be a solution to {\sc lex-prob}.

For example, in downsampled lexicase selection \citep{hernandez_random_2019, ferguson_characterizing_2020}, there is a parameter, $D$, which can be any value between 1 and the number of fitness criteria ($M$). On every generation, $M*\frac{1}{D}$ fitness criteria are randomly selected. Only these fitness criteria are used on that generation. One can trivially reduce any instance of lexicase selection to an instance of downsampled lexicase selection by setting $D=1$. This same basic approach works for truncated lexicase selection \citep{spector_relaxations_2018}, MADCAP lexicase selection \citep{spector_relaxations_2018}, lexicase selection with weight shuffle \citep{troise_lexicase_2018}, and Lexi$^2$ \citep{de_lima_lexi_2022}. 

Of course, some lexicase selection variants exist under which calculating $P_{lex}$ is not actually $NP$-Hard. For example, fixed-order lexicase-based tournament selection (FOLBaT) \citep{burks_investigation_2018, troise_lexicase_2018} determines a fixed ordering of fitness criteria to use. Selecting the order takes polynomial time, and then, because it does not change, $P_{lex}$ can be calculated in polynomial time because the (exponentially many) other orderings do not need to be considered.

\section{An optimized brute-force algorithm for {\sc lex-prob}}

As {\sc lex-prob} is $NP$-Hard, we should not expect to find a polynomial time solution to it in the near future. However, it is still possible to develop a much faster algorithm than the naive brute-force approach presented in \citep{la_cava_probabilistic_2018}. While the time complexity of this algorithm will still be exponential, we can dramatically improve the problem sizes that it can handle in reasonable amounts of time. Doing so yields a practically meaningful improvement in the cases where $P_{lex}$ can be calculated.

Using the recursive approach presented in \citep{la_cava_probabilistic_2018} as a starting point, the following polynomial-time optimizations can be made at every level of recursion:

\begin{enumerate}
    \item Eliminate all candidate solutions that are strictly worse than all other remaining solutions on the fitness criteria still under consideration (\textit{i.e.} those which cannot possibly be selected).
    \item Eliminate all fitness criteria on which all remaining candidate solutions are tied (\textit{i.e.} those which cannot possibly have an effect on the outcome of selection).
    \item Rather than continuing recursion when only two candidate solutions remain, just calculate the probability of each one winning. 
\end{enumerate}

These optimizations dramatically increase the cases where calculating $P_{lex}$ is tractable. A highly-optimized C++ implementation (and accompanying Python wrapper) are available at in the supplemental information for this paper.

\section{Conclusion}

I have proved that calculating the probabilities of an individual solution being selected under lexicase selection or most of its variants is $NP$-Hard. This result is relevant to anyone interested in calculating these probabilities in polynomial time. Doing so would be as hard as finding a polynomial time solution to many other well-known difficult problems that no one has found a polynomial time solution to despite decades of effort. Indeed, it may very well be impossible. As such, this is an important result both for researchers studying the theory of lexicase selection and for those studying practical ways to improve lexicase selection.

This proof also has interesting implications about the relationship between $\epsilon$-lexicase selection and standard lexicase solution. While it was obvious that instances of lexicase selection can be trivially converted to instances of $\epsilon$-lexicase selection, the fact that the conversion can be done in the opposite direction is more surprising. This finding suggests that the core innovation of  $\epsilon$-lexicase selection is that it enables various different subsets of solutions to tie on a given criterion. Moreover, it suggests that, if it were for some reason necessary to recreate the behavior of $\epsilon$-lexicase selection using standard lexicase selection (\textit{e.g.} if for some reason floating point numbers needed to be avoided), doing so would likely be possible.

Finally, by tying these problems into the network with $NP$-Hard problems, these results allow us to leverage discoveries about those problems and apply them back to lexicase selection. In particular, many solvers for $NP$-Hard problems have been written that are very fast in practice even if their worst-case run-time is still non-polynomial. Additionally, much work has been done on approximation algorithms for $NP$-Hard problems. In cases where calculating $P_{lex}$ efficiently is necessary, this body of knowledge can now be leveraged.

\begin{acks}
I would like to thank Josh Nahum and members of the Michigan State University ECODE lab for their helpful comments on an early draft of this paper.
\end{acks}

\bibliographystyle{ACM-Reference-Format}
\bibliography{main}










\end{document}